\documentclass[preprint,review,12pt]{elsarticle}
\usepackage{amssymb}
\usepackage{amsmath}
\usepackage{array}
\usepackage{booktabs}
\usepackage{xcolor}

\usepackage{graphicx} 
\usepackage{multirow}
\usepackage{fullpage}
\usepackage{tabularx}
\usepackage{rotating, makecell}
\usepackage{svg}
\usepackage{algpseudocode}
\usepackage{algorithmicx}
\usepackage{algorithm}

\begin{document}
\begin{frontmatter}

\title {Visual Geo-Localization from images} 
 \author{Rania SAOUD}
\author{Slimane LARABI}
\ead{slarabi@usthb.dz}
\affiliation{organization={Computer Science Faculty, USTHB University}, addressline={BP 32}, 
            city={El Alia},
            postcode={16111}, 
            state={Algiers},
            country={Algeria}}
\begin{abstract}
This paper presents a visual geo-localization system capable of determining the geographic locations of places (buildings and road intersections) from images without relying on GPS data. Our approach integrates three primary methods: Scale-Invariant Feature Transform (SIFT) for place recognition, traditional image processing for identifying road junction types, and deep learning using the VGG16 model for classifying road junctions. The most effective techniques have been integrated into an offline mobile application, enhancing accessibility for users requiring reliable location information in GPS-denied environments. 
\end{abstract}


\end{frontmatter}

\section{Introduction}

Geo-localization involves determining the precise location of an image or video in the real world, typically using latitude and longitude coordinates obtained through technologies such as GPS, Wi-Fi, and IP address tracking. Algorithms process this data to pinpoint exact coordinates\cite{11}\cite{12}. Geo-localization is important for organizing and analyzing large volumes of imagery data, as demonstrated by systems like the US Geological Survey (USGS), which classify and locate satellite and drone images to streamline data collection and analysis. Social media platforms like Instagram use geo-localization to tag photos with specific locations, enabling users to explore location-based content\cite{11}.
Despite its significance, many images and videos lack geo-localization data, particularly those collected in the past or by devices without GPS capabilities\cite{12}. This presents challenges in organizing and analyzing such images based on their locations. Visual geo-localization addresses these issues by identifying the geographic position of an image using visual cues within the image itself, rather than relying on external metadata like GPS tags\cite{12}. As defined by Brejcha and Cadik, visual geo-localization involves finding the geographic coordinates (and possibly the camera orientation) for a given query image\cite{12}. This technique uses attributes such as building facades, points of interest, and geographical information to compare the query image with a database of geo-referenced images\cite{11}. The goal is to provide accurate location predictions in situations where conventional methods, such as GPS, are ineffective, inaccurate, or unavailable.
Advanced computer vision and deep learning algorithms play a critical role in visual geo-localization by analyzing visual data to identify unique features or patterns associated with specific locations. This technology enhances applications like augmented reality, self-driving cars, and location-based services. However, geo-localization faces technical challenges due to the vast amount of visual content shared online. Issues such as ambiguity from geographical features appearing in multiple locations and the need for precision and accuracy in geotagging processes arise. Additionally, many images lack geo-tags, complicating initial location estimation. The dynamic nature of physical surroundings, influenced by factors like construction and seasonal changes, affects the visual indicators used for geo-localization. Poor photographic conditions, such as inadequate lighting and camera distortions, further complicate accurate location identification\cite{11}.
To address these challenges, scalable systems capable of processing and matching large quantities of data with extensive geographic databases are necessary. Such systems must achieve high performance and accuracy to be useful in applications like city navigation or emergency services, where pinpointing locations within a few meters is important. Efficiency is essential for real-time applications, which depend on quick data processing. Robustness against various conditions and low impact from domain shift, where query images significantly deviate from reference data, are also critical\cite{13}.
These requirements underscore the need for innovative methods combining advanced deep learning and computer vision techniques, with a strong emphasis on user privacy, to analyze visual information and accurately predict geographic locations.
This paper introduces a novel geo-localization solution developed at the University of Science and Technology Houari Boumediene. To enhance the system's efficiency, we have collected data from various locations over time. Our project explores three methods for visual geo-localization: using SIFT, identifying road junctions through image processing, and applying deep learning techniques. After evaluating each method, we selected the most effective approach. This method has been developed into an offline mobile app, providing users with convenient and reliable location identification without the need for an internet connection.

\section{Related Works}
The field of image-based geo-localization has seen various innovative approaches aimed at improving the accuracy of determining geographic locations from images. For instance, the method presented in \cite{1} uses sequences of ground photos compared against geo-tagged aerial images to handle changes in perspectives and sequence differences effectively. Similarly, the TransVLAD: Multi-Scale Attention-Based Global Descriptors for Visual Geo-Localization \cite{2} introduces an efficient technique that significantly enhances location accuracy, which is especially useful for autonomous driving and robot navigation.
Urban localization accuracy is improved in \cite{3} by focusing on buildings and using advanced matching techniques to handle different viewing conditions and perspectives. Another novel approach is presented in \cite{4}, which mimics human navigational methods to efficiently determine locations from photos. In \cite{5}, a sophisticated method enhances the accuracy of photo localization by ensuring that visual features match well with those in a large database.
Cross-view Image Geo-localization \cite{6} introduces a streamlined model for matching street-level photos with aerial images, simplifying the process and enhancing accessibility. The study in \cite{7} shows how selectively using image features for matching can improve the efficiency and accuracy of localization, introducing a robust method for feature encoding.
Large-scale Image Geo-Localization Using Dominant Sets \cite{8} uses an advanced clustering approach to select the most coherent and compact sets of features matching the query image, significantly speeding up the localization process. The work in \cite{9}, RK-Net, addresses the challenge of localizing images from different viewpoints by focusing on the most informative parts of images, simplifying the learning process and enhancing performance.
Finally, Semantically Guided Location Recognition for Outdoor Scenes \cite{10} enhances the accuracy of outdoor scene localization by focusing on man-made structures through semantic segmentation, proving effective in environments with significant visual variability.
Together, these studies underscore the diversity of strategies used to address the complexities of image-based geo-localization, each contributing uniquely to advancements in the field.

\section{Proposed System}

Our system aims to enhance geo-localization accuracy using visual information from images. By integrating advanced techniques in image processing, machine learning, and computer vision, we can pinpoint geographic locations independently of GPS data. We use three principal approaches: the Scale-Invariant Feature Transform (SIFT), deep learning with the VGG16 model, and traditional image processing methods for analyzing road junction types.
Instead of using traditional image processing techniques to identify types of road junctions  directly from image, road junctions and its time-consuming nature, we used a deep learning approach using the VGG16 model. In addition, we used the SIFT approach to match images with the model images.
This section details the steps of our proposed visual geo-localization solution:
We start with an existing map that provides a basic layout of the geographical area of interest. This map serves as the reference for aligning and comparing our image-based findings. The coordinates of the starting position are obtained from the name of the starting place.

\textbf{Using SIFT for Place Recognition:}\\
We use the Scale-Invariant Feature Transform (SIFT) to analyze panoramic images \cite{18}. This method identifies the closest match to a given location by comparing features from new images with those in our panoramic dataset.

\textbf{Deep Learning for Junction Identification:}\\
In this advanced method, we use the VGG16 deep learning model to classify road junctions from images. Once we identify these junctions, we match them with the map to verify and enhance geographical accuracy.

\subsection{Place Recognition SIFT descriptor}

In this approach, we focus on place recognition by comparing input images against a dataset of panoramic images using SIFT descriptors. First, we create the dataset by extracting frames from videos, which are preferred over still images due to their continuous and comprehensive environmental capture. This continuity is important for seamless stitching, so we ensure a significant overlap—typically around 30-50\%—between consecutive frames. Videos must be recorded with stable movement and consistent speed. After creating panoramic views, we refine them by cropping and applying the Scale-Invariant Feature Transform (SIFT) technique to extract distinctive features. Subsequently, we use the Detectron2's panoptic segmentation to filter the input images, retaining only those where buildings and their surroundings occupy a significant portion of the content (more than 60\%). We then apply the SIFT algorithm to these selected images to match their features against the panoramic dataset using the FLANN matcher. This step identifies the most similar place for each image. Finally, a voting process determines the most likely place based on the results from all analyzed images.

\textbf{1) Dataset Creation :}\\

The initial step in our dataset creation involves extracting frames from video recordings at specific intervals determined by the skip\_frames parameter. This method balances the comprehensiveness of the data against redundancy, ensuring the frames are representative yet manageable in size. To construct panoramic images, we stitch multiple frames together using algorithms that align overlapping areas based on shared features. This process helps overcome challenges such as changes in lighting and camera movement, and we enhance the accuracy of the stitch through techniques like color normalization and feature alignment. 

\begin{figure}[ht!]
    \centering
    \includegraphics[width=12cm]{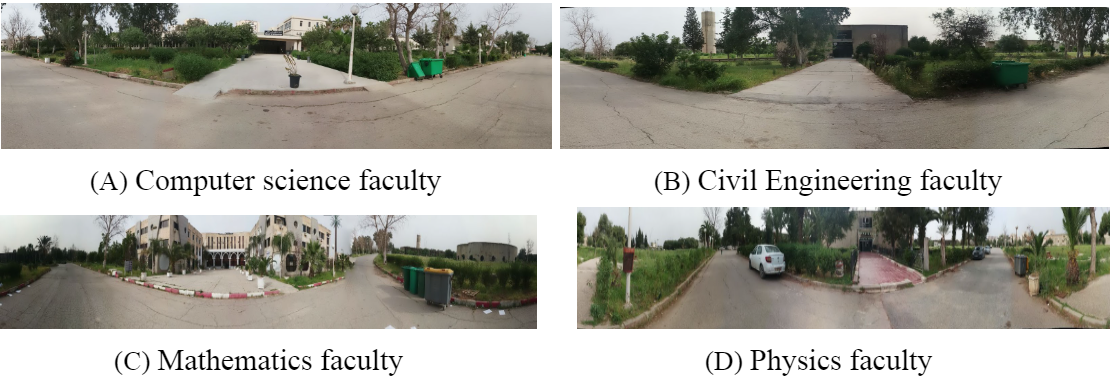}
    \caption{Panoramic dataset taken in our university}
    \label{fig1}
\end{figure}

After stitching, we crop unnecessary black edges from the panoramic images to focus on relevant content, thereby optimizing the data for further processing. Using the SIFT technique, we then extract key features from the images that remain consistent despite changes in scale or rotation, making it easier to match images from different scenes. These features are saved in .npz files for efficient use across various computing platforms. Additionally, we use APIs like Google Places to gather detailed metadata about specific locations, enriching our dataset with precise geographical coordinates. These coordinates are integrated using Geographic Information Systems (GIS) for detailed mapping and spatial analysis, enhancing the practical application of our geolocation tasks. Figure \ref{fig1} represents a panoramic dataset taken in our university. 

\textbf{2) Image Classification and Filtering using Detectron2 :}\\

\begin{figure}[ht!]
    \centering
    \includegraphics[width=12cm]{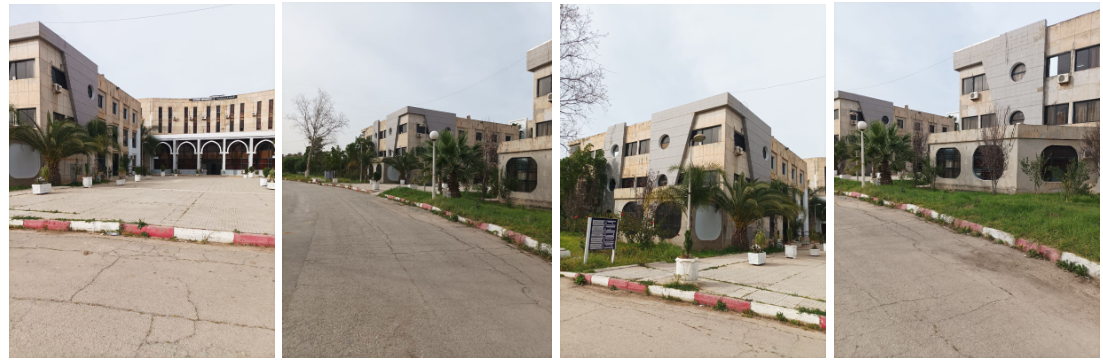}
    \caption{Example of the selected images for place matching using detectron2 classification}
    \label{fig2}
\end{figure}

Detectron2, a tool created by Facebook AI, performs panoptic segmentation on the images provided by the user. The model uses Panoptic FPN R-101 from Detectron2’s model zoo. Each image is first adjusted to the RGB color format, which is necessary for processing with Detectron2. The model then segments the images, identifying different elements like roads and pavements. It calculates how much of each image is covered by roads and pavements. A threshold is set so that only images where roads and pavements make up less than 40\% of the image move on to the next phase. This ensures that only images with significant non-road elements are chosen for place matching using SIFT. For instance, Figure \ref{fig2} displays examples of images selected for further processing.

\textbf{3) Place Matching using SIFT : }\\

\begin{figure}[ht!]
    \centering
    \includegraphics[width=8cm]{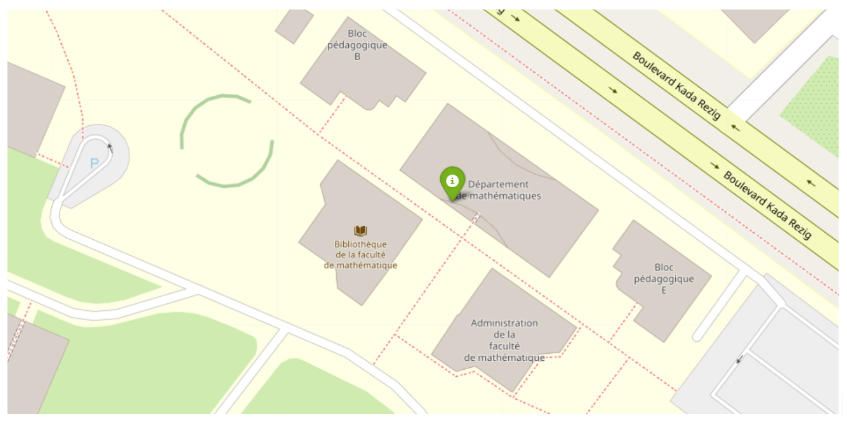}
    \caption{Visual representation of the most similar place}
    \label{fig3}
\end{figure}

We take the images classified in the previous step for SIFT matching and generate SIFT descriptors for each image. We use the FLANN-based matcher for its efficiency in handling large datasets. The matcher compares the SIFT descriptors of the query image to those in the dataset using the k-nearest neighbors approach. For each query image, the system calculates the number of 'good matches'—those matches that pass a distance test indicating high similarity. This is done by checking if the distance of the nearest match is less than 75\% of the distance of the second nearest match, thus eliminating less likely matches. Each query image contributes a 'vote' to the place it matches best based on the count of good matches, so the place receiving the highest number of votes across all queries is determined to be the most similar place. Finally, we generate a visual representation using Folium by placing a green marker on a map at the coordinates corresponding to the most voted place, providing an intuitive geographical context to the results. Figure \ref{fig3} is an example of the most similar place for the mathematics faculty building images.

\subsection{Road junction Identification using deep learning}

The second approach is to identify road junctions using a deep learning method with the VGG16 model, pre-trained on a large database of images. We start by collecting diverse images of roads, including features like T-junctions, X-junctions, Y-junctions, and roundabouts. These images are labeled to highlight specific road features. The VGG16 model is then fine-tuned to focus on road junctions, ensuring it performs well with new, unseen images. After training, the model is tested with new images to evaluate its capability in identifying and classifying different road features.

\textbf{Model Preparation and Training:}

- Preprocessing: Images were resized to 224x224 pixels and normalized to suit the input requirements of the VGG16 model.

- Class Weight Calculation: To address dataset imbalances, class weights were calculated and applied during training to ensure equitable learning across less frequent classes.

- Model Architecture Modifications: The base VGG16 model was customized by removing the top layer and adding dense layers with ReLU activation and dropout layers to enhance feature learning capabilities.

\textbf{Model Training and Validation:}

- Training Set Preparation:The dataset was split into training, validation, and test sets, with augmentation techniques such as random flipping and brightness adjustment applied to enhance generalization.

- Training Process:  Initially, only the top layers of the 
model were trained. Techniques like batch normalization and dropout were used to improve training stability and prevent over-fitting.

\textbf{Fine-Tuning and Evaluation:}

- Fine-Tuning: After the initial training, deeper layers of the VGG16 were gradually unfrozen and the model was fine-tuned with a reduced learning rate to refine its ability to classify junction types accurately.

- Model Evaluation: The fine-tuned model was tested against a separate set of images to assess its performance and ensure it could reliably classify road features in varied real-world conditions.

Our deep learning strategy harnesses the capabilities of the pretrained VGG16 model, which we fine-tune to classify various road junctions and features.

\textbf{Advantages:}

Generalization: Once trained, the model can effectively identify a broad spectrum of road features, adapting seamlessly to new, encountered scenarios.

Efficiency: Compared to real-time computer vision analysis, deep learning models require less computational power during inference, making them faster and more cost-effective.

Ease of Integration: These models can be effortlessly integrated into various applications due to their flexibility and self-contained nature.

\textbf{Challenges:}

Data Dependency: The precision of predictions heavily depends on the quantity and quality of the training data.

Opacity in Decision-Making: Unlike rule-based techniques, deep learning models do not easily reveal how decisions are made, which can complicate troubleshooting and refinement.

Considering the limitations of the Computer Vision approach, particularly its dependency on specific imaging conditions and the complexity of adapting to diverse scenarios, we transitioned to using a Deep Learning approach. This change significantly enhances our system's ability to generalize and scale. By expanding our dataset and continuously retraining the model, we can accommodate a wider array of road features and intersections without the need for extensive manual algorithm adjustments. This strategic pivot not only speeds up the process but also broadens the potential applications of our model in real-world scenarios.

\subsection{Integration and Map Matching}

This section outlines the methodology for integrating geographic data with a graph-based map system and the subsequent application for route mapping and validation.

1) Creation of Graph-based Road Map Dataset

Defining the Area of Interest with JOSM: Initially, we define the geographic area of interest using JOSM, a robust tool for editing OpenStreetMap data. This tool enables us to tailor the map data precisely by adding or removing roads, buildings, and intersections, thus facilitating the map-matching process.

\begin{figure}[ht!]
    \centering
    \includegraphics[width=8cm]{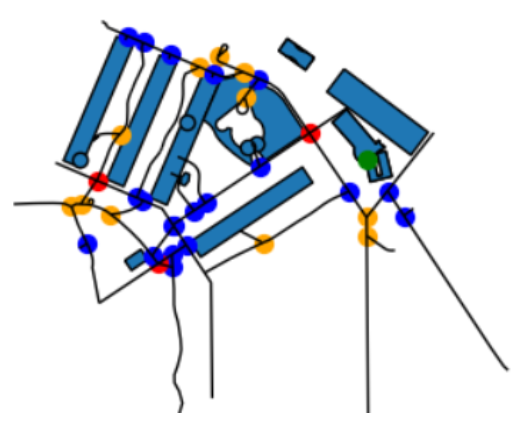}
    \caption{Junction types: X-junctions (red), Y-junctions (yellow), T-junctions (Blue) and crossroads}
    \label{fig4}
\end{figure}

Automated Road Feature Analysis and Classification: This process involves converting detailed road data into a structured graph. We use GeoJSON files to represent roads as line strings, which aids in constructing a graph where intersections are nodes and road segments are edges, enhancing our analysis of road connectivity. We classify junction types by identifying nodes where roads intersect, with different configurations indicating various junction types like T-junctions, X-junctions or Y-junctions (see Figure \ref{fig4}).

Integration and Validation: Integrating the data into a unified graph-based map involves consistency checks to ensure data accuracy and representation. Validation occurs through systematic verification of node adjacency's and junction classifications, ensuring the integrity of the geographic data within our application.

2) Sequence Matching with Graph-Based Map

In this phase, we use the graph-based road map to identify potential routes by matching a sequence of junction types to the graph data. Starting with intersections close to a given location, we calculate distances using the Haversine formula to maintain focus on relevant intersections. We then apply a depth-first search algorithm to trace paths that match the defined sequence of junction types. This method not only aids in route planning but also enhances our navigation system's reliability by ensuring the paths adhere to expected road patterns.

\begin{figure}[ht!]
    \centering
    \includegraphics[width=12cm]{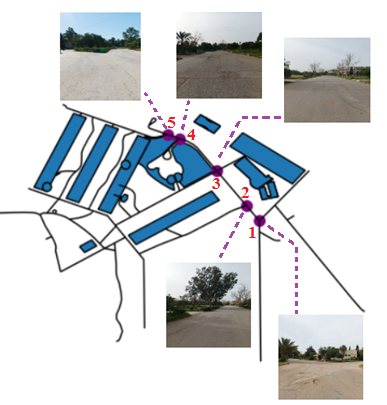}
    \caption{ The images represent the road intersections encountered along the route. These include various junctions such as T-junctions, Y-junctions, and crossroads. The images provide a visual representation of the real-world path.The path computed by our algorithm is indicated by purple dots which correspond to the real path images.}
    \label{fig5}
\end{figure}

Example of path finding result :  

In this section, we present an example of the pathfinding results generated by our algorithm. Figure \ref{fig5}  illustrates a sequence of real images taken from intersections 1 to 5. Each image represents a critical point in the route, showcasing different types of road junctions encountered along the way.

The pathfinding algorithm processes these images to determine the most probable route on the map. By using the intersection types and their sequences, the algorithm identifies the path through the road network. The resultant path is depicted in the map shown in Figure \ref{fig5}.

3) Validation and Correction Mechanism

Once a route is identified, we conduct a detailed validation of each junction of the sequence to confirm alignment with the expected path. Discrepancies are addressed by exploring alternative routes that provide a better match. This process is important for maintaining the accuracy and reliability of the navigation system, ensuring that the route suggestions are practical and reflect real-world conditions.

\begin{algorithm}
\caption{Sequence Matching with Graph-Based Map}
\begin{algorithmic}[1]
\State \textbf{Input:} Coordinates $(lon, lat)$, Road type sequence
\State \textbf{Data:} Intersections dataset

\Procedure{main}{}
    \State Load dataset into DataFrame
    \State Find closest intersections within radius
    \For{each intersection}
        \If{road type matches first in sequence}
            \State Find path from this intersection
            \If{path found}
                \State \textbf{break}
            \EndIf
        \EndIf
    \EndFor
    \State Plot the path
\EndProcedure

\Function{find\_path}{$current\_node, sequence, df$}
    \If{sequence is empty}
        \State \Return path
    \EndIf
    \For{each adjacent node}
        \If{node matches next type in sequence}
            \State Continue finding path
        \EndIf
    \EndFor
    \State \Return None
\EndFunction

\end{algorithmic}
\end{algorithm}

\section{Experimental results}

In this section, we present the results of our experiments conducted to evaluate the performance of our visual Geo-localization system. We will detail the performance metrics for each approach, compare the effectiveness of traditional image processing and deep learning methods, and discuss the results of our mobile application implementation.

\subsection{Dataset Preparation and Annotation}

\begin{figure}[ht!]
    \centering
    \includegraphics[width=10cm]{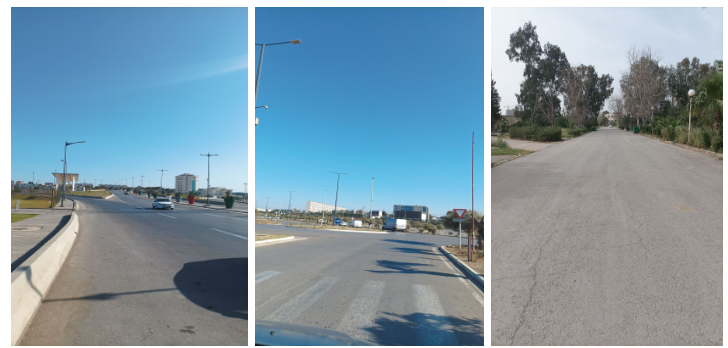}
    \caption{Few collected images.}
    \label{fig6}
\end{figure}

We captured 282 images of road junctions using a smartphone, focusing on T-junctions, Y-junctions, X-junctions, and roundabouts. These images were taken from various distances and angles, ensuring a diverse dataset,a selection of images is displayed in Figure \ref{fig6}.

\begin{figure}[ht!]
    \centering
    \includegraphics[width=8cm]{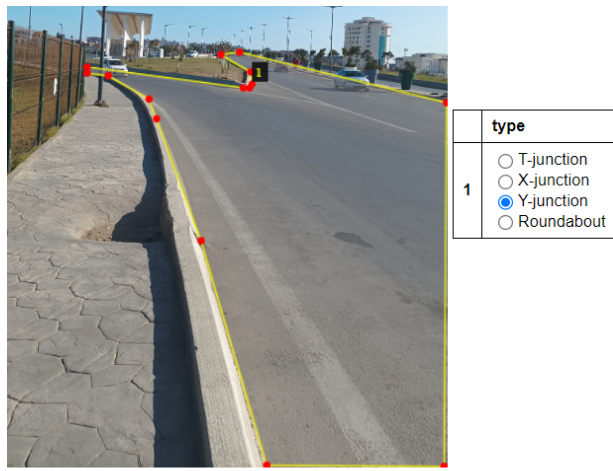}
    \caption{Example of image annotation}
    \label{fig7}
\end{figure}

Using the VGG Image Annotator (VIA), we labeled the images to highlight specific road features. Accurate labeling is important for the model's performance, as shown in Figure \ref{fig7}.

\subsection{Data Splitting and Balancing}
The data was split into training (70\%), validation (15\%), and test (15\%) sets. To address class imbalances, we used SMOTE (Synthetic Minority Over-sampling Technique) to generate synthetic samples.

\subsubsection{Data Augmentation}

To enhance the model's generalization, we used Keras' ImageDataGenerator for data augmentation. This technique artificially creates variations of our existing images by applying random transformations such as shifting, rotating, flipping, and altering brightness. These transformations increase the diversity of the dataset, helping the model to generalize better to new, unseen images.
Additionally, SMOTE was used to create synthetic samples for the minority classes. SMOTE works by generating new samples along the line segments joining existing minority class samples. This further balances the dataset and ensures that all classes are equally represented during training.
Combining ImageDataGenerator and SMOTE significantly increased the number of training images. While the exact number of augmented images is not directly counted, the process resulted in more than 20,000 image variations over the training epochs, greatly enhancing the dataset's diversity.
\subsection{Training and Fine-Tuning of the VGG16 Model}

We used the Adam optimizer and categorical cross-entropy loss function for training the VGG16 model. The accuracy metric was used to measure the model's performance during training.
The initial training was conducted over 50 epochs with a batch size of 32. For fine-tuning, we unfroze the last eight layers of the VGG16 model, training for an additional 30 epochs with a reduced learning rate to improve accuracy.
We used callbacks like ReduceLROnPlateau and EarlyStopping to enhance training efficiency and prevent over-fitting.

\subsection{Results}
\begin{figure}[ht!]
    \centering
    \includegraphics[width=12cm]{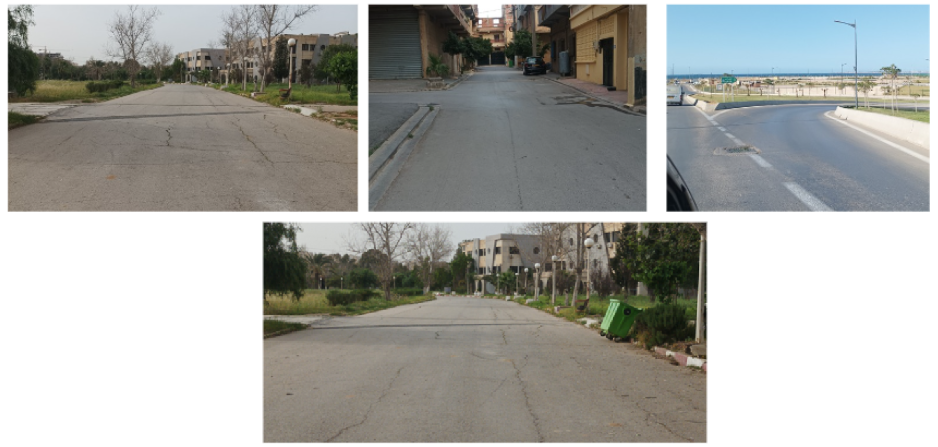}
    \caption{(Top) Sample of images, well classified with respectively X, T and Y junction. (Bottom) Sample of a misclassified image, the correct junction is “X”}
    \label{fig8}
\end{figure}

We conducted tests where users uploaded images of buildings and road intersections. The app processed these images to determine the user's location, prioritizing intersection images for better accuracy. Figure \ref{fig8} shows an example of three query images with a correct classification. Table 01 the values of precision, recall and F1-measure.

\begin{table}[ht!]
    \centering
    \label{tab1}
    \begin{tabular}{|c|c|c|c|}
    \hline
         & Precision & Recall & F1-score\\
    \hline
        Roundabouts & 1.00 & 0.67 & 0.8\\
    \hline
        T-junction & 1.00 & 1.00 & 1.00\\
    \hline
        X-junction & 1.00 & 1.00 & 1.00\\
    \hline
        Y-junction & 0.95 & 1.00 & 0.97\\
    \hline
        Accuracy & & & 0.98\\
    \hline
\end{tabular}
\caption{Performance metrics of our fine-tuned VGG16 model.}
\end{table}

By addressing the imbalance in the dataset, the model's performance on the test set improved by 3\%. Below, the table 02 illustrates the changes in loss and accuracy.

\begin{table}[ht!]
    \centering
    \label{tab1}
    \begin{tabular}{|c|c|c|}
    \hline
         & Model Before Class Weights & Model After Class Weights \\
    \hline
        Training Loss & 0.04 & 0.0294 \\
    \hline
        Training Accuracy & 0.97 & 0.9846 \\
    \hline
        Test Loss & 0.53 & 0.0959 \\
    \hline
        Test Accuracy & 0.94 & 0.9767 \\
    \hline
\end{tabular}
\caption{Training and test loss and validation accuracy's before and after addressing the imbalance in the dataset using SMOTE.}
\end{table}

\section{Conclusion}
The experimental results demonstrate the effectiveness of our proposed Geo-localization method using a deep learning approach. The proposed system which is implemented on mobile successfully integrates these methods, providing a practical tool for users in GPS-denied environments. Future work will focus on expanding the dataset, further improving model accuracy, and adapting the system for augmented reality (AR) and indoor applications.

\end{document}